\documentclass[10pt,twocolumn,letterpaper]{article}

\usepackage{iccv}
\usepackage{times}
\usepackage{epsfig}
\usepackage{graphicx}
\usepackage{amsmath}
\usepackage{amssymb}

\usepackage{booktabs}
\usepackage{bbding}
\usepackage{pifont}%
\usepackage[pagebackref=true,breaklinks=true,colorlinks,bookmarks=false]{hyperref}

\usepackage[accsupp]{axessibility} 

\usepackage[breaklinks=true,bookmarks=false]{hyperref}

\iccvfinalcopy 

\usepackage[capitalize]{cleveref}
\crefname{section}{Sec.}{Secs.}
\Crefname{section}{Section}{Sections}
\Crefname{table}{Table}{Tables}
\crefname{table}{Tab.}{Tabs.}

\def\iccvPaperID{2927} 

\newcommand{\cmark}{\ding{51}}%
\newcommand{\xmark}{\ding{55}}%

\ificcvfinal\fi

\begin{document}

\title{Randomized Quantization: A Generic Augmentation  \\ for Data Agnostic Self-supervised Learning}




\author{Huimin Wu$^{1}\thanks{Equal contribution. Work done during an internship at MSRA.}$  \quad\quad Chenyang Lei$^{2}$\footnotemark[1] \quad\quad Xiao Sun$^{4}$ \quad\quad Peng-Shuai Wang$^{3}$  \\ Qifeng Chen$^{1}$  \quad\quad Kwang-Ting Cheng$^{1}$ \quad\quad Stephen Lin$^{5}$ \quad\quad Zhirong Wu$^{5}$ \vspace{4pt} \\
    $^1$HKUST  \quad
    $^2$CAIR, HKISI\_CAS \quad
    $^3$Peking University \quad
    $^4$Shanghai AI Lab \quad
    $^5$Microsoft Research Asia\\
}

\maketitle
\ificcvfinal\thispagestyle{empty}\fi

\begin{abstract}

Self-supervised representation learning follows a paradigm of withholding some part of the data and tasking the network to predict it from the remaining part.
Among many techniques, data augmentation lies at the core for creating the information gap.
Towards this end, masking has emerged as a generic and powerful tool where content is withheld along the sequential dimension, e.g., spatial in images, temporal in audio, and syntactic in language.
In this paper, we explore the orthogonal channel dimension for generic data augmentation by exploiting precision redundancy.
The data for each channel is quantized through a non-uniform quantizer, with the quantized value sampled randomly within randomly sampled quantization bins.
From another perspective, quantization is analogous to channel-wise masking, as it removes the information within each bin, but preserves the information across bins.
Our approach significantly surpasses existing generic data augmentation methods, while showing on par performance against modality-specific augmentations.
We comprehensively evaluate our approach on vision, audio, 3D point clouds, as well as the DABS benchmark which is comprised of various data modalities.
The code is available at~\url{https://github.com/microsoft/random_quantize}.



\end{abstract}

\section{Introduction}

We are witnessing a convergence of multi-modal AI~\cite{devlin2018bert,bao2021beit} where the architecture and the learning algorithm are unified for various data modalities. 
This exciting direction abandons the domain-specific knowledge for an individual data modality, but rather pursues a solution far more generalizable. 

For self-supervised representation learning, masked modeling~\cite{devlin2018bert} or simply masking input as an augmentation~\cite{wu2022extreme} has emerged as an effective approach. The input data is represented by a 2D tensor with a sequential dimension and a channel dimension in a modality-agnostic way~\cite{baevski2022data2vec}. The sequential dimension can be spatial in images, temporal in audio, and syntactic in languages. 
The masking mechanism withholds information along the sequential dimension, and exploits it for supervision. As a result, models learned from the masking supervision demonstrate strong capability for capturing correlations between sequential tokens~\cite{he2022masked}.
\begin{figure}
  \centering
  \includegraphics[width=1\linewidth]{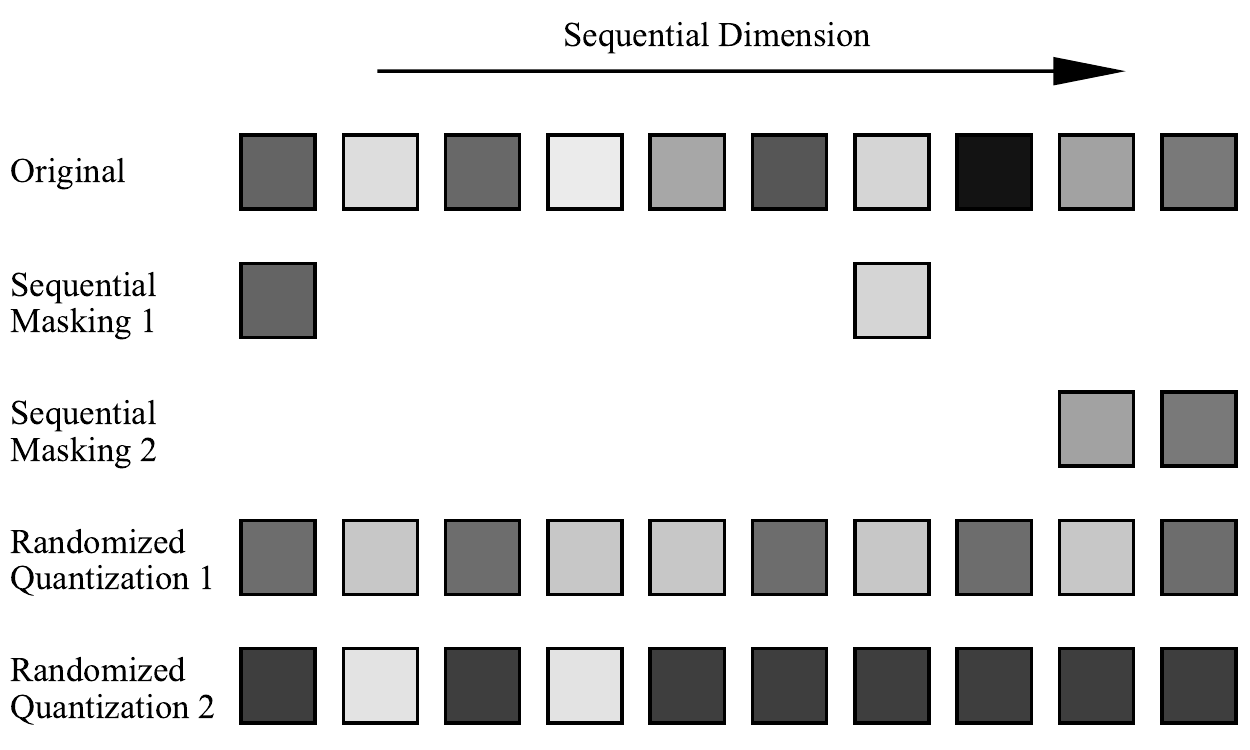}
  \caption{
  We represent data as a matrix with a sequential dimension and a channel dimension. As a generic data augmentation, masking drops tokens along the sequential dimension. The proposed randomized quantization instead withholds information along the channel dimension. In this figure, we use 1D data of 10 sequential tokens for illustration. Data values are coded in grayscale.}
  \label{fig:short}
\end{figure}

The channel dimension describes the data feature at each sequential location, for example, RGB color at a spatial location or spectrogram frequency at a time step.
Despite being generic, masking approaches have neglected to exploit supervision along the channel dimension.
While the number of channels for images is as small as three, the channels for audio and tabular data can be as many as hundreds. 
Formulating the self-supervision from the channel dimension holds much potential for representation learning.

In this paper, we draw a connection between the sequential masking operation and quantization by exploring quantization as a novel form of masking along the channel dimension.
The data in each channel is dynamically quantized through a non-uniform quantizer, with the quantization value randomly sampled from randomly sampled quantization bins.
In this way, information within each quantization bin is masked out, yet information across bins is retained. 
The information removed by quantization is controlled by the number of bins and the size of the bins, which has been rigorously studied in theory~\cite{shannon1959coding}. 
The larger the distortion rate, the stronger the quantization when it is used as an augmentation for representation learning. 
The extreme case of using only a single bin is equivalent to dropping the entire channel.
We systematically study various quantization configurations for their effects as a data augmentation, for example, with respect to the number bins, uniform or non-uniform bins, and methods to select quantization values.

We apply the randomized quantizer as the only augmentation, or in conjunction with augmentations along the sequential dimension on state-of-the-art Siamese representation learning methods 
MoCo-v3~\cite{chen2021empirical} and BYOL~\cite{grill2020bootstrap}. In comparisons with domain-agnostic augmentations based on MixUp~\cite{zhang2017mixup}, our approach achieves state-of-the-art results by a large margin on vision, audio, 3d point cloud, and the DABS benchmark.
Compared with domain-specific augmentations, our approach achieves competitive performance against handcrafted augmentations on vision, and state-of-the-art performance on audio and 3d point clouds.


Our contributions can be summarized as follows:
\begin{itemize}
\vspace{-5pt}
    \item[-] We propose a simple yet effective data-agnostic augmentation for contrastive learning, based on quantization along the channel dimension.
    \item[-] We design a randomization technique  by varying the quantization value and the quantization bins, to enhance data augmentation.
    \item[-] We demonstrate the generality and strong performance of our channel-wise randomized quantization for vision, audio, 3D point clouds and the DABS benchmark in a data-agnostic way.
\end{itemize}


\section{Related Works}

\begin{figure*}
  \centering
  \includegraphics[width=1\linewidth]{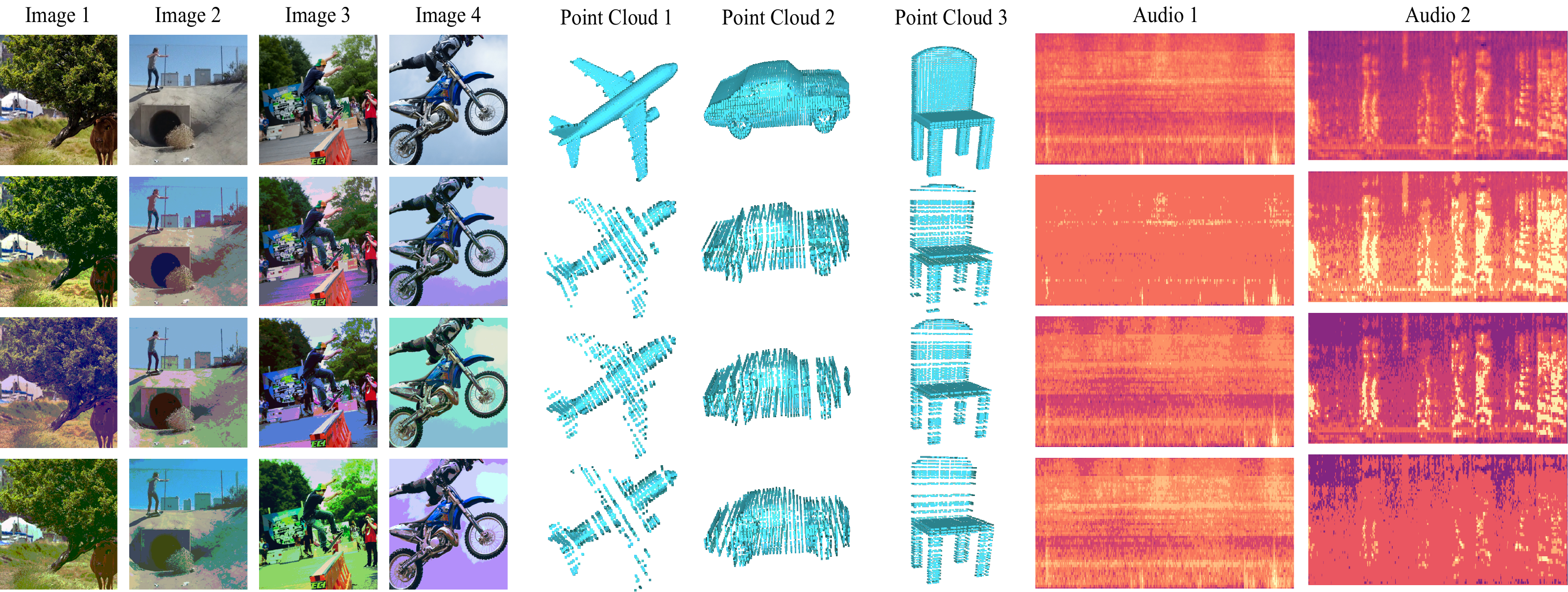}
  \caption{Visualizing randomized quantization augmentation on images, 3d point clouds, and audio. The first row presents the original signal, and the bottom three rows are augmented views. Randomized quantization alters color and enhances edges on images, spatially samples coordinates on point clouds, and enhances frequency channels for audio.}
  \label{fig:vis_img_audio}
\end{figure*}

\begin{figure}
  \centering
  \includegraphics[width=1\linewidth]{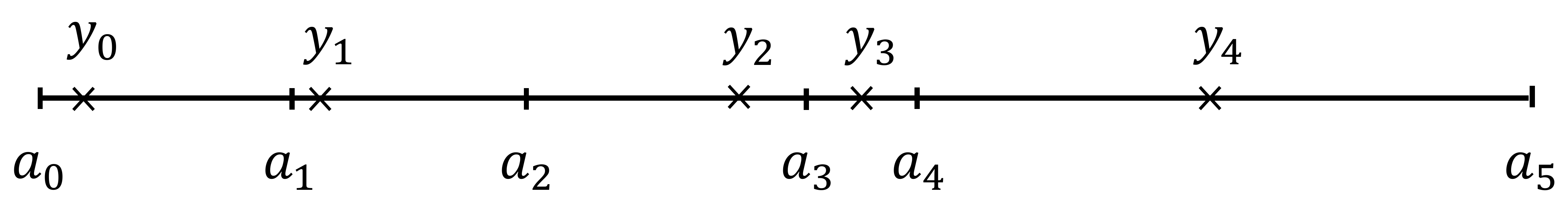}
  \caption{Illustration of a non-uniform quantizer with five bins. $a_i$ denotes the interval of the quantization bin and $y_i$ denotes the reproduction value from each bin.}
  \label{fig:quantizer}
  \vspace{-3pt}
\end{figure}

\noindent
\textbf{Self-supervised learning} extracts labels from the data itself and tasks the network to learn transferable representations. 
Among the earliest forms of self-supervised models are auto-encoders~\cite{hinton1993autoencoders} and generative models~\cite{hinton2009deep}. But since the input and the output are identical, a neural network may easily find shortcuts and use memorization to solve the generation task.
Advances in recent years show that information needs to be withheld from the input to prevent cheating~\cite{doersch2015unsupervised}. 
Pretext tasks such as colorization~\cite{zhang2016colorful}, inpainting~\cite{pathak2016context}, jigsaw puzzles~\cite{noroozi2016unsupervised} were proposed in vision, while masked modeling~\cite{devlin2018bert}, next sentence prediction~\cite{kiros2015skip,jernite2017discourse}, and replaced word prediction~\cite{clark2020electra} were developed in natural language processing.
Speediness~\cite{benaim2020speednet,huang2021ascnet} and temporal order~\cite{wei2018learning,misra2016shuffle} have been exploited for video representation learning.
Due to space limitations, we omit the literature for speech~\cite{baevski2020wav2vec}, tabular data~\cite{arik2021tabnet}, graph-structured data~\cite{sun2019infograph} and many other modalities. 
The optimal pretext task for each target problem may be different. However, there exists enormous interest in obtaining a single foundation model~\cite{bommasani2021opportunities} for all downstream applications.

Instead of withholding data for supervision, contrastive models~\cite{wu2018unsupervised,oord2018representation} create new data via data augmentation and compare features extracted using a Siamese network for supervision.
Siamese representation learning can be categorized by whether to use explicit negatives~\cite{grill2020bootstrap}, ways to define negatives~\cite{bachman2019learning}, and various loss formulations~\cite{caron2021emerging,zbontar2021barlow}. However, the main driving signal for learning lies in the  augmentations.

\vspace{2pt}
\noindent
\textbf{Data augmentation} enlarges the number of data instances by leveraging prior knowledge of the data and target problem properties. For supervised learning, data augmentation aids in reducing overfitting and regularization~\cite{zhang2021understanding}. 
For self-supervised learning, the information gap created by two augmentations provides learning supervision.
Typically, the data augmentation function extracts partial information from the data and optionally adds corruptions.

Popular image-specific augmentations include cropping, scaling, color jittering, Gaussian blurring, cut-out~\cite{devries2017improved}, cut-mix~\cite{yun2019cutmix}, and auto-augment, which searches for a data augmentation policy~\cite{cubuk2018autoaugment}. In natural language processing, synonym replacement~\cite{wei2019eda}, back translation~\cite{brislin1970back}, random word insertion and deletion are most common. For audio and speech, altering the pitch, changing the playback speed, and masking either along the time axis or the frequency axis~\cite{park2019specaugment} may improve performance.
Additionally, augmenting data through a generative model~\cite{bowles2018gan} such as a GAN is a viable approach.


\vspace{2pt}
\noindent
\textbf{Domain-agnostic augmentation} aims to generalize modality-specific and domain-specific augmentations into a unified formulation. Finding such general priors in data is challenging. One line of work follows Mixup~\cite{zhang2017mixup}, which is initially proposed to improve empirical risk minimization of supervised learning by linearly interpolating data and labels. 
Because of its generality, later works have explored its application on other data modalities~\cite{guo2020nonlinear}, a wide range of problems~\cite{lucas2018mixed,hendrycks2019augmix}, as well as representation learning~\cite{tamkin2021dabs,lee2020mix, verma2021towards}.
Another important line of work generalizes masked modeling~\cite{devlin2018bert}, which was initially proposed for language modeling, to other data modalities and domains~\cite{he2022masked,tong2022videomae,xu2022masked}.
The masking mechanism samples a subset of the input data, while Mixup introduces additional corruptions which are not observed in the original data instance.
A third line of work tries to mine inherent regularities and structures in the data. 
NNCLR~\cite{dwibedi2021little} and MSF~\cite{koohpayegani2021mean} use nearest neighbors derived from the current representations as a novel form of augmentation for contrastive learning.
Our randomized quantization differs from these works, and augments data along the channel dimension.


\vspace{2pt}
\noindent
\textbf{Quantization} represents numerical values with a fixed discrete set of numbers so as to reduce communication bandwidth and maintain representation quality.  
The rounding error was first analyzed a century ago~\cite{sheppard1897calculation}, 
and the theory based on variable-rate quantization~\cite{shannon1948mathematical} and Huffman coding~\cite{huffman1952method} revolutionized the communications industry. We refer readers to a survey~\cite{gray1998quantization} that describes this area from a theoretical perspective. 

Quantization for efficient neural networks~\cite{gholami2021survey} aims to reduce neural network latency while maintaining model accuracy. The advances of half-precision~\cite{banner2018scalable,wang2018training} and mixed-precision training~\cite{courbariaux2014training,gupta2015deep,micikevicius2017mixed} has accelerated model training by an order of magnitude. Works have shown that neural networks can be completely binarized~\cite{lin2017towards,wu2015adjustable,courbariaux2015binaryconnect} with reasonable performance. 
Stochastic quantization~\cite{chen2020statistical,fan2020training,bengio2013estimating} is a technique for learning and compressing model weights in a way that avoids local minima with the low-precision weight representations.

There exist two works~\cite{cao2022synergistic,fu2022contrastive} that study the use of quantization in contrastive representation learning. In both, quantization is applied on the intermediate features and model weights. Though effective for model compression, the quantization of features/weights can be viewed as unsuitable for augmentation
due to the unpredictability of the corresponding change in the input.
Our work additionally differs from these methods~\cite{cao2022synergistic,fu2022contrastive} in its randomization of a non-uniform quantizer. This technique creates a much more complex augmentation space than uniform quantization, generating valid images of greater diversity.

\section{Approach}

\begin{figure*}[t]
  \centering
  \includegraphics[width=1\linewidth]{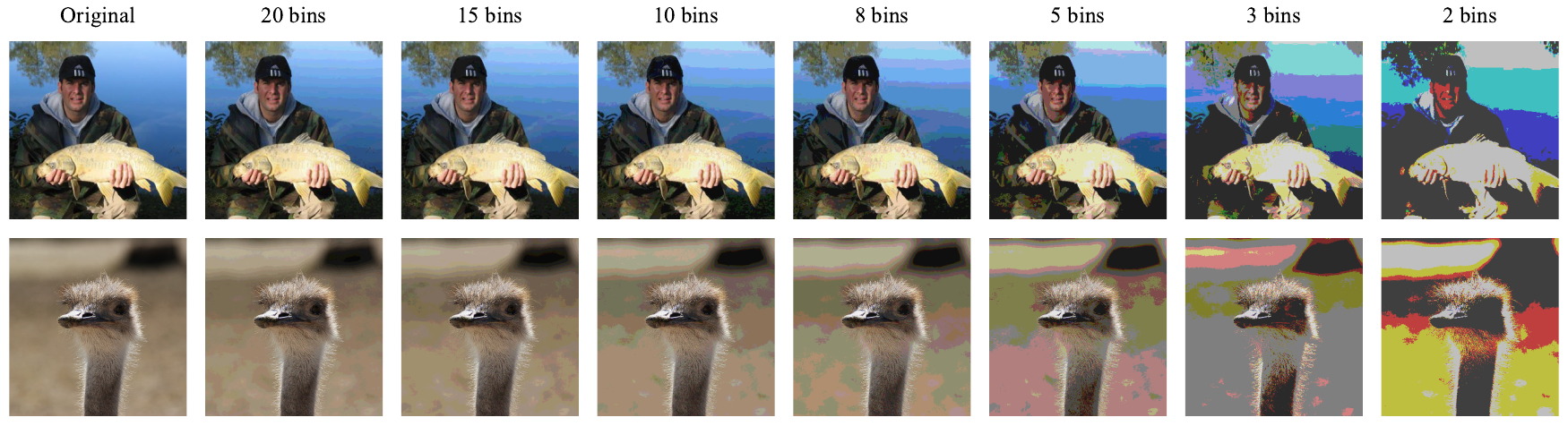}
  \caption{Visualization of quantized images with different numbers of bins. The images are quantized by a uniform quantizer. Fewer than three quantization bins causes severe information reduction, while fifteen or more bins leads to negligible difference from the original image. An intermediate number of bins (e.g., five to ten) is well-suited for image augmentation.}
  \label{fig:img_bins}
\end{figure*}

This paper proposes a novel generic data augmentation for representation learning based on quantization.
We first provide preliminaries on quantization. We then introduce two factors to inject randomness into the quantization procedure. 

\subsection{Preliminaries: Quantization}

A quantizer is a function which consists of a set of non-overlapping intervals or bins $S = \{S_i=[a_i, a_{i+1}))\}, i=0,1,...n-1$, and a set of reproduction values $y_i$. $n$ is the number of intervals and reproduction values. Let $x$ be a one-dimensional input signal. The quantizer maps values within an interval to a single scalar, defined as $q(x) = y_i$ for $ x\in S_i$. Formally, it can be written as
\begin{equation}
    q(x) = \sum_i y_i \cdot 1_{S_i}(x),
\end{equation}
where the indicator function $1_{S_i}(x) = 1$ if $x\in S_i$ and $1_{S_i}(x) = 0$ otherwise. Figure~\ref{fig:quantizer} gives an illustration of a quantizer with five intervals.
Quantization represents the original signal using a finite number of bits and hence introduces error to signal recovery. The central research problem is to find better tradeoffs between communication bandwidths and reproduction errors.

Quantization can be categorized by uniform quantization and non-uniform quantization. A uniform quantizer has intervals and values which are evenly spaced, whereas a non-uniform quantizer allows either intervals or values to be unevenly spaced. 
Uniform quantizers are amenable to hardware deployment. However, non-uniform quantizers may perform better depending on the probabilistic distribution of $x$.

\begin{figure}
  \centering
  \includegraphics[width=1\linewidth]{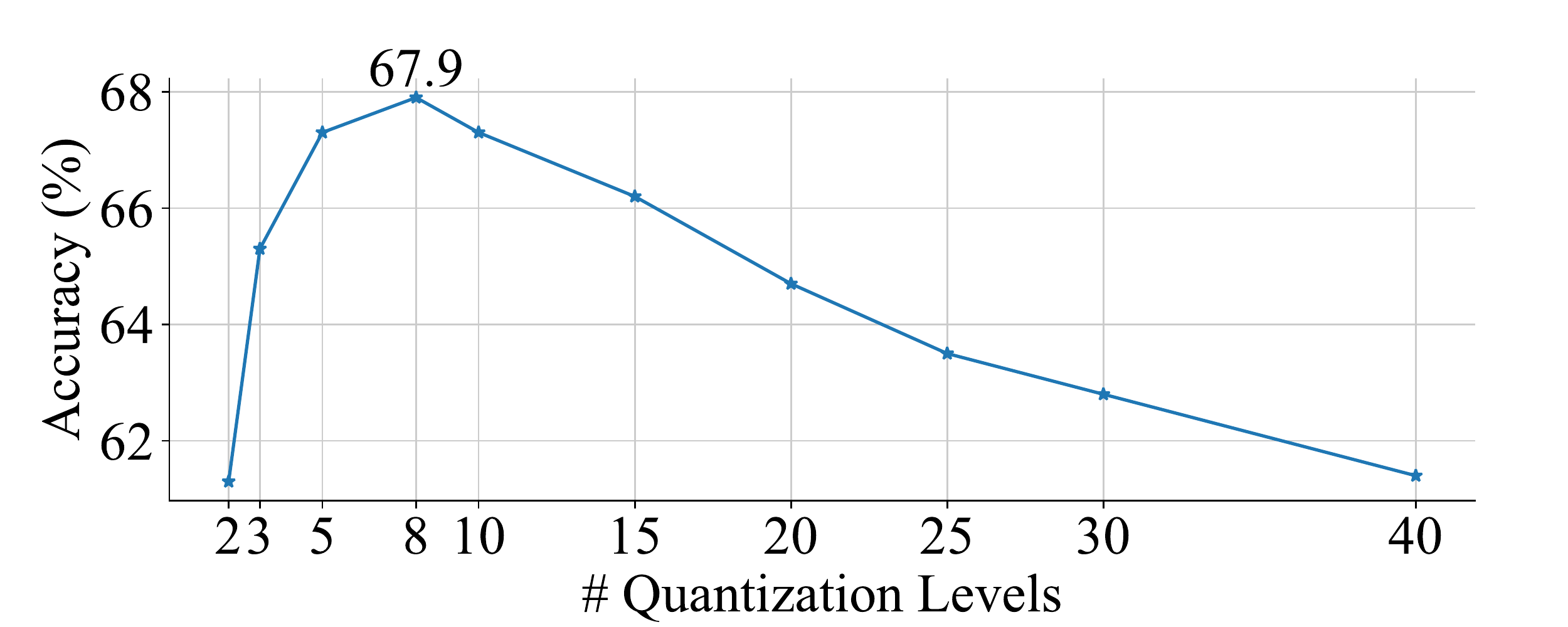}
  \caption{Ablation study on the number of quantization bins. The peak performance is reached at 8 bins. Fewer bins deliver heavier augmentations and larger bins lead to weaker augmentations.}
  \label{fig:numbins}
\end{figure}

\subsection{Randomized Quantization as Augmentation}

We aim to exploit quantization as a data withholding tool for representation learning. The information within each quantization bin is withheld, while the information across bins are retained. 
For data augmentation, a key aspect is the complexity of the augmentation space. We design a complex quantization augmentation by randomizing the intervals and the reproduction values. Concretely, given $S_i = [a_i, a_{i+1})$, $a_i$ is generated by
\begin{equation}
    a_0, a_1, ... ,a_{n-1} = \text{sort}(a'_0, a'_1, ..., a'_{n-1}) 
\end{equation}
\begin{equation}
\label{eq:random1}
   a'_i = U(\text{min}(x), \text{max}(x)), i=0,1,...,n-1,
\end{equation}
where $U$ denotes random sampling with a uniform distribution over the interval, and $\mathrm{min}(x)$/$\mathrm{max}(x)$ refers to the minimum/maximum of the values $x$ per channel.
The reproduction value $y_i$ is randomly sampled within the corresponding interval,
\begin{equation}
\label{eq:random2}
    y_i = U (a_i, a_{i+1}).
\end{equation}
The resultant randomized quantizer is non-uniform. The number of quantization bins $n$ is the hyperparameter of the augmentation. 

\begin{table}[t]
\centering

\begin{center}
\setlength{\tabcolsep}{3.4pt}
\begin{tabular}{c|c|c|c}
& random bins & random values & top1 acc\\
\hline
baseline & \xmark & \xmark & 50.0 \\
+ quantize & \xmark & \xmark & 54.8 \\
+ quantize  &\xmark & \cmark & 62.6 \\
+ quantize & \cmark & \xmark & 66.0 \\
+ quantize & \cmark & \cmark &\textbf{67.9}\\
\end{tabular}
\end{center}
\caption{ 
Ablation study of the two randomness factors for randomized quantization described in Eq.~\ref{eq:random1} and Eq.~\ref{eq:random2}.  We examine the effect of randomized bins and random reproduction values for each bin. These two factors increase the complexity of the augmentation and significantly improve the performance.
}
\label{tab:abl_non_uniform}
\end{table}

\subsection{Data-Agnostic Augmentation}

The proposed randomized quantization augmentation can be applied to the channel dimension for any arbitrary data modality. The physical interpretation of the augmentation depends on the nature of the data modality. In Figure~\ref{fig:vis_img_audio}, we visualize the augmentations for images, audio and point clouds.
On images, it removes the high frequency details but highlights object boundaries and edges. It also alters color appearance significantly.
On audio, we examine the augmented sound acoustically and we find the augmentation tends to enhance specific frequency signals, such as low-frequency sounds or high-frequency sounds.
On point clouds where the channel dimension represents xyz coordinates,  it tends to downsample local structures but highlight the global shape.
Some discrete modalities such as language are not directly amenable to quantization, but can be mapped to a continuous representation via a frozen data embedding layer. In Section~\ref{exp_dabs}, we examine randomized quantization on such input embeddings.

\subsection{Siamese Representation Learning}

Siamese representation learning or contrastive learning relies heavily on the quality of the augmentations~\cite{grill2020bootstrap,zhao2021distilling}.
We apply the proposed augmentation on Siamese representation learning. At each training iteration, we sample two views from a data instance using randomized quantization by itself or in conjunction with other augmentations. 
Loss terms such as InfoNCE~\cite{oord2018representation} and L2 are applied on the two views. The MoCo-v3~\cite{chen2021empirical} and BYOL~\cite{grill2020bootstrap} frameworks are followed in this paper, and we refer readers to the original papers for details.

\section{Ablation Study}

\begin{table}[t]

\begin{center}
\begin{tabular}{c|c|c|c}
&100-ep&300-ep&800-ep\\
\hline
MoCo-v3 &67.9&71.6& 72.1\\
BYOL & 67.2&71.0 & 71.6\\
\end{tabular}
\end{center}
\caption{ 
Representation learning with randomized quantization augmentation benefits from more training epochs without early saturation.} 
\label{tab:training_length}
\end{table}

\begin{table}[t]

\begin{center}

\setlength{\tabcolsep}{12pt}
\begin{tabular}{l|c|c}

%
Augmentations&MoCo-v3&BYOL\\
\hline
CR&10.1&9.9\\
CR + DACL~\cite{verma2021towards}&32.7&33.2\\
CR + i-Mix~\cite{lee2020mix}&30.3&28.7\\
CR + SSQL~\cite{cao2022synergistic} & 11.1& 6.4\\
CR + Ours&\textbf{42.9}& \textbf{43.0}\\

\hline

RRC&50.0&49.3\\
RRC + DACL~\cite{verma2021towards} &57.2&57.6\\
RRC + i-Mix~\cite{lee2020mix}& 55.4&49.9\\
RRC + SSQL~\cite{cao2022synergistic} &42.9 & 49.3\\
RRC + Ours&\textbf{67.9}&\textbf{67.2}\\

\end{tabular}
\end{center}
\caption{Comparisons with alternative domain-agnostic augmentation techniques under the linear classification protocol on ImageNet. CR is short for center crop, and RRC is short for random resized crop. Our randomized quantization approach achieves the best results against prior arts.}
\label{tab:cmp_daa}
\end{table}

We choose visual representation learning for an ablation study. 
Random resized cropping is taken as the baseline augmentation, and we apply our randomized quantization after it.
Following the MoCo-v3 framework~\cite{chen2021empirical}, we use ResNet-50~\cite{he2016deep} as the backbone network.
The optimizer is consistent with MoCo-v3, and the network is optimized for 100 epochs.
Representation learning is conducted on the ImageNet-1K dataset~\cite{deng2009imagenet}, and linear classification accuracy is reported on the validation set.

We ablate three design factors of the proposed quantization-based augmentation which affect its ability to mask channel-wise information. 

\vspace{2pt}
\noindent
\textbf{Randomizing Bins.}
The performance of representation learning depends on the complexity of the pretext tasks created from random augmentations.
In Table~\ref{tab:abl_non_uniform}, the baseline approach using the random resized crop augmentation obtains 50.0\% top-1 accuracy. 
Using a fixed uniform quantizer improves the performance mildly to 54.8\%. 
Randomizing the locations and sizes of bins allows for uneven masking and creates more effective pretext tasks. It improves the performance significantly to 66.0\%.


\begin{table}[t]
\begin{center}

\setlength{\tabcolsep}{12pt}
\begin{tabular}{l|c|c}
Method& MoCo-v3 & BYOL\\
\hline
Ours & 42.9 & 43.0 \\
RRC & 50.0 & 49.4\\
RRC + CJ & 60.1 & 61.1\\
RRC + Ours & 67.9 & 67.2\\
Full & \textbf{68.9} & \textbf{68.9} \\

\end{tabular}
\end{center}
\caption{ 
Comparisons with image-specific augmentations under the linear classification protocol on ImageNet. CJ stands for color jittering, and Full includes random resized crop, color jittering, grayscaling, Gaussian blurring and solarization. Randomized quantization is stronger than color jittering by a large margin. It falls behind the full handcrafted augmentations by just 1\%.
}
\label{tab:cmp_specific}
\end{table}

\vspace{2pt}
\noindent
\textbf{Randomizing reproduction values.}
Quantization is also affected by how each bin is represented.
Commonly, the values within a bin's range are represented by the midpoint. As an alternative, we also consider taking a random value in the range.
Intuitively, random reproduction values lead to bias in the quantization error, making them no longer zero-mean and bringing a stronger augmentation effect. 
It is found to benefit representation learning, yielding an increase of 1.9 points upon randomizing the bins in Table~\ref{tab:abl_non_uniform}.

\vspace{2pt}
\noindent
\textbf{Number of quantization bins.} 
Figure~\ref{fig:numbins} illustrates the effect of various numbers of quantization bins. 
Intuitively, fewer bins leads to a stronger masking effect and higher cross-view variation.
We vary the number of bins and find strong performance with 5-10 bins, peaking at 8 bins. 
This observation is similar to spatial masking in MAE~\cite{he2022masked} where an optimal masking ratio is found. 
In Figure~\ref{fig:img_bins}, we visualize quantized images with different numbers of bins. To make the visualization consistent and comparable, we use the uniform quantizer in this case.
It can be observed that too much information is withheld when using too few bins, and too many bins withholds too little information.



\vspace{2pt}
\noindent
\textbf{Training epochs.} 
We further study training the augmentations with more epochs. 
In Table~\ref{tab:training_length}, the performance improves from 67.9\% with 100 epochs to 71.6\% with 300 epochs and 72.1\% with 800 epochs. With this complex augmentation, the network benefits from longer training.


\section{Experiments on Various Modalities}

We examine pre-training with the proposed augmentation across a variety of modalities including 1) vision (Section~\ref{exp_images});  2) 3D point clouds (Section~\ref{exp_point_cld}); 3) audio (Section~\ref{exp_audio}); and 4) the DABS benchmark~\cite{tamkin2021dabs} (Section~\ref{exp_dabs}) comprised of data from multiple domains.
The hyper-parameter $n$ indicating the number of bins is tuned for each modality. 
We leave the description of corresponding datasets, settings and evaluation metrics to each section.

\subsection{Images}
\label{exp_images}

\begin{figure}[t]
  \centering
  \includegraphics[width=1\linewidth]{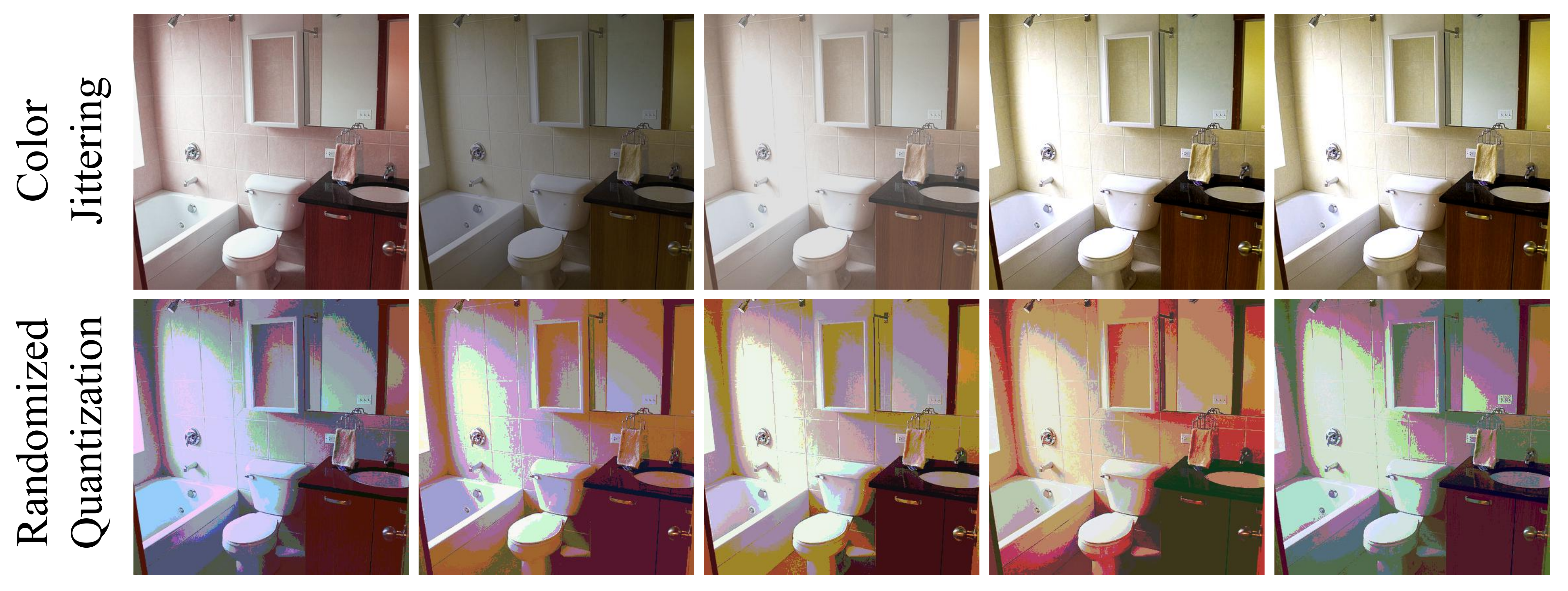}
  \caption{Visual comparisons between color jittering and randomized quantization. Randomized quantization exhibits greater change in visual appearance and stronger edge enhancement.}
  \label{fig:img_cmp_color_jittering}
\end{figure}

We compare the proposed randomized quantization augmentation against domain-agnostic augmentation baselines, as well as domain-specific augmentations designed for images. The number of quantization bins is chosen as $n=8$.
The experimental protocol follows the ablation study. 

\vspace{2pt}
\noindent
\textbf{Comparisons with domain-agnostic augmentations.}
Recent works on domain-agnostic augmentation are predominantly adapted from Mixup~\cite{zhang2017mixup}.
For example, i-Mix~\cite{lee2020mix} linearly interpolates input data, and their corresponding virtual labels are generated from the current batch.
Similarly, DACL~\cite{verma2021towards} interpolates input but uses it as a way of adding noise to the original data.
We also consider a baseline SSQL~\cite{cao2022synergistic} which jointly performs quantization and contrastive learning for lower bit-width deployment.

In Table~\ref{tab:cmp_daa}, we compare our approach to these methods on two spatial operations: center crop (CR) and random resize crop (RRC). Center crop amounts to no augmentation, and random resized crop is frequently used in vision applications.
Our evaluation is based on two Siamese representation learning frameworks MoCo-v3 and BYOL, since BYOL is said to have different behavior on augmentations. 

Randomized quantization performs the best against Mixup-based augmentations. 
As a standalone augmentation, randomized quantization obtains an accuracy of 42.9\% with MoCo-v3, which outperforms DACL and i-Mix by a large margin.
In conjunction with random resized crop, a 10\% margin is maintained. 
SSQL heavily relies on domain-specific augmentations, and in many cases, it fails to improve the baseline when handcrafted augmentations are removed.
The results using MoCo-v3 and BYOL training objectives are similar. Overall, randomized quantization achieves state-of-the-art results against domain-agnostic baselines in the vision domain.


\begin{table}[t]
\begin{center}
\begin{tabular}[0.1\textwidth]{l|c|c|c}
 Method & Backbone & Epochs & Top1 accuracy\\
 \hline
 MSF w/w & R-50 &200 & 66.3\\
 \hline
 Ours& R-50 & 100 & \underline{67.9}\\
 Ours& R-50 & 300 & \textbf{71.6}\\
 \end{tabular}
\end{center}
\caption{Comparison with weakly-augmented SSL baseline MSF.} 
\label{tab:cmp_weak_aug_ssl}
\end{table}

\begin{table}[t]
\begin{center}
\setlength{\tabcolsep}{3pt}
\begin{tabular}{l|c|c|c|c|c|c}
&1\%&2\%&5\%&10\%&20\%&100\%\\
\hline
FoldingNet (lin) &56.4 &66.9& 75.6& 81.2& 83.6&88.4\\
MID-FC (lin)&61.5 &73.1& \textbf{80.2}& 84.2& {86.9}&90.3\\
Ours (lin)& \textbf{66.7} &	\textbf{74.3}&80.0&\textbf{84.5}&	\textbf{87.2} &\textbf{90.5}\\
\hline
Scratch &58.5& 71.2& 80.1& 85.4& 88.7&92.9\\
MID-FC (ft)&67.3 &76.5& 83.6& {88.4}& 90.2&\textbf{93.0}\\
Ours (ft)&\textbf{71.3}&	\textbf{78.5}&	\textbf{84.9}&	\textbf{88.6}&	\textbf{90.6}&\textbf{93.0}\\

\end{tabular}
\end{center}
\caption{Linear probing (lin) and finetuning (ft) results for the shape classification task on the ModelNet40 dataset. 
Pre-training is conducted on the ShapeNet dataset.
Our augmentation improves the classification accuracy substantially on various ratios of data, especially on very limited (1\%) data. 
}
\label{tab:pointcls}
\end{table}

\begin{table}[t]
\begin{center}



\setlength{\tabcolsep}{2.8pt}
\begin{tabular}{l|c|c|c|c|c|c}
&\multicolumn{3}{c|}{C.mIoU}&\multicolumn{3}{c}{I.mIoU}\\
\cline{2-7}
&1\%&5\%& 100\%&1\%&5\%& 100\%\\
\hline
Multi-Task (lin) & - & 73.9 &- & 68.2 & 80.7 & -  \\
MID-FC (lin)&66.2& 76.5&82.8&72.4&80.9&84.1\\
Ours (lin)&\textbf{70.6}&\textbf{76.9}&\textbf{82.9}&\textbf{77.4}&\textbf{81.9}&\textbf{84.3}\\
\hline
MID-FC (ft)&67.6&77.8&84.3&76.2&82.1&\textbf{85.5}\\
Ours (ft)&\textbf{69.5}&\textbf{78.4}&\textbf{
84.4}&\textbf{77.8}&\textbf{82.3}&\textbf{85.5} \\
\end{tabular}
\end{center}
\caption{Linear probing (lin) and finetuning (ft) results for the shape segmentation task on the ShapeNet Part dataset. Pre-training is conducted on ShapeNet. Our augmentation improves the performance substantially on various ratios.
} 
\label{tab:pointseg}
\end{table}



\begin{table*}[t]
\begin{center}

\setlength{\tabcolsep}{3pt}
\begin{tabular}{c|c|c|c|c}
Name&Task&\#Classes &Data size& Avg duration (s)\\
\hline
NSynth (NS)~\cite{engel2017neural}&Musical instrument classification&11&305,979&4.0\\
UrbanSound8K (US8K)~\cite{salamon2014dataset}&Urban sound classification &10&8,732&4.0\\
VoxCeleb1 (VC1)~\cite{nagrani2017voxceleb}&Speaker identification&1,211&153,514&8.2\\
VoxForge (VF)~[from~Voxforge.org]&Language identification&6&176,438&5.8\\
Speech Commands V2 (SPCV2)~\cite{warden2018speech}&Command
classification &35&105,829&1.0\\
Speech Commands V2 (SPCV2/12)~\cite{warden2018speech}&Command
classification &12&105,829&1.0\\

\end{tabular}
\end{center}
\caption{Downstream dataset details for audio representation learning.}
\label{tab:audio_datasets}
\end{table*}

\begin{table*}[t]
\begin{center}

{
\setlength{\tabcolsep}{12pt}
\begin{tabular}{c|cccccc|c}
Method&NS&US8K&VC1&VF&SPCV2/12&SPCV2&Average \\
\hline
TRILL~\cite{shor2020towards}&  - &-&17.9&88.1&74.9&-&-\\
COLA~\cite{saeed2021contrastive}&63.4&-&29.9&71.3&71.7&62.4&-\\
OpenL3~\cite{cramer2019look}&-&78.2&-&-&-&-&-\\
COALA~\cite{favory2020coala}&73.1&72.7&-&-&-&-&-\\
COLA~\cite{saeed2021contrastive}&70.2&78.5&30.4&79.5&76.7&76.8&68.7\\
BYOL-A~\cite{niizumi2021byol} &74.1&79.1&40.1&90.2&91.0&92.2&77.8\\
\hline
Ours&\textbf{74.2}&78.0&\textbf{45.7}&\textbf{92.6}&\textbf{95.1}&92.1&\textbf{79.6}\\

\end{tabular}
}
\end{center}
\caption{Linear probing results for audio representation learning on six downstream datasets. Pre-training is conducted on the AudioSet dataset. Our model outperforms BYOL-A on four of the six datasets, with an average improvement of 1.8\%.}
\label{tab:cmp_audio}
\end{table*}

\begin{table*}[ht]
\begin{center}


\setlength{\tabcolsep}{10pt}
\begin{tabular}{c|cccccc|c}
Method&Natural Images&Text&Speech&Sensors&Chest x-rays &Images \& Text & Average \\
\cline{1-8}
Scratch&10.1&42.3&24.9&69.8&68.1&\textbf{57.5} & 45.5 \\
e-Mix&27.9&44.1&41.8&79.5&72.4&48.9 & 52.4\\
Ours&\textbf{32.1}&\textbf{44.7}&\textbf{44.5}&\textbf{84.9}&\textbf{73.4}&54.5&\textbf{55.6}\\
\end{tabular}
\end{center}
\caption{
Evaluation of the representation performance over six modalities in the DABS benchmark. Representations are trained on a single primary dataset for each modality and evaluated on a number of downstream datasets. The performance for each modality is averaged across the downstream datasets and shown in the table.
}
\label{tab:cmp_dabs}
\end{table*}

\vspace{2pt}
\noindent
\textbf{Comparisons with domain-specific augmentations.}
We further compare with image-specific augmentations for visual representation learning in Table~\ref{tab:cmp_specific}. 
We find that randomized quantization is much stronger than color jittering, which is heavily designed with prior knowledge such as brightness, contrast, and saturation for pixels.  
In Figure~\ref{fig:img_cmp_color_jittering}, we visualize color jittering and our augmentation. It can be observed that our augmentation leads to stronger and more diverse visual appearances than color jittering. 
Our augmentation is 1\% weaker than the full augmentation, which includes random resized crop, color jittering, grayscaling, Gaussian blurring and solarization successively. 




\noindent\textbf{Comparison with weakly-augmented SSL baselines.}
We compare the performance with MSF~\cite{koohpayegani2021mean} in a weak augmentation setting.
MSF implicitly utilizes close nearest neighbors as a form of augmentation.
 In Table~\ref{tab:cmp_weak_aug_ssl}, our method is better, and it benefits from longer training (67.9 vs. 71.9).
Additionally, we demonstrate our approach on diverse modalities, while MSF only focuses on vision.

\subsection{3D Point Clouds}
\label{exp_point_cld}
We explore self-supervised representation learning on point clouds, represented by a disordered set of xyz coordinates.
The pretraining is conducted on the ShapeNet~\cite{chang2015shapenet} dataset consisting of 57,449 3D shapes.
Octree-based Sparse CNN~\cite{wang2017cnn} is used as the backbone network, which takes 3D point clouds as input and extracts point features as well as shape features.
We follow the MID-Net~\cite{wang2021unsupervised} model as the baseline, which is trained by a point-wise and instance-wise contrastive loss.
The model is trained by a SGD optimizer with a batch size of 32 and a weight decay of 5e-4. The initial learning rate is 0.03 and decreases by a factor of 10 after 200 and 300 epochs, and the training process terminates after 400 epochs.
We apply the randomized quantization augmentation after the base augmentations used in MID-Net. 
Unlike images and audio which are snapped to grids, strong quantization of point cloud coordinates drastically degrades 3d point data. 
We thus choose to use a larger number of bins, n=30, in order to maintain more information. In practice, since the 3d points are sparsified by quantization, we observe a substantial training speedup as a side benefit. 
For evaluation, we experiment on two downstream tasks: shape classification and segmentation.

Shape classification is conducted on ModelNet40~\cite{wu20153d} which is composed of 13,834 3D models from 40 categories.
    For each shape, we extract a global feature with the pre-trained backbone then train a linear classifier, or finetune the network, and report the average classification accuracy in Table~\ref{tab:pointcls}. 
We do comparison with FoldingNet~\cite{Yang2018a} and MID-FC~\cite{wang2021unsupervised}.
With our augmentation,  we improve the classification accuracy  over the baseline MID-FC~\cite{wang2021unsupervised} substantially, especially when the training data is limited as shown in Table~\ref{tab:pointcls}.
For example, with 1\% of the training data, we improve the classification accuracy by 5.2 and 4.0 points on linear probing and finetuning, respectively.

Shape segmentation is conducted on ShapeNet Part~\cite{yi2017large} with 16,881 3D point clouds from 16 categories.
For each shape, we extract point-wise features with the pre-trained backbone then train a segmentation head composed of two fully connected layers, or finetune the network, and report the mean IoU across all categories (C.mIoU) and the mean IoU across all instances (I.mIoU) in Table~\ref{tab:pointseg}.
We do comparison with two unsupervised pretraining methods: MID-FC~\cite{wang2021unsupervised} and Multi-Task~\cite{Hassani2019}.
Our results are consistently better than the baselines across different ratios of training data.
And similarly to ModelNet40 classification, we observe significant improvements with limited (1\% and 5\%) training data.
For example, with 1\% of the training
data, our method improves the segmentation performance by 4.4 and 1.9 IoU on linear probing and finetuning, respectively.


\subsection{Audio}
\label{exp_audio}
We apply randomized quantization to audio representation learning.
We use AudioSet~\cite{gemmeke2017audio} as the pretraining dataset, with 1,963,807 audio samples of 527 classes.
The pretrained representation is evaluated on six downstream audio classification datasets, summarized in Table~\ref{tab:audio_datasets}.

We largely follow the experimental settings of BYOL-A~\cite{niizumi2021byol} and treat it as our baseline. 

We convert audio clips into the commonly used log-scaled spectrogram representation. 
Random resized crop is used to extract a $64\times96$ frequency-temporal segment for training. 
We replaced the Mixup augmentation used in BYOL-A with our randomized quantization, with the number of bins set to 5.
We follow prior works~\cite{niizumi2021byol} by using a lightweight 2D convolutional network as the backbone.
We train the network using the Adam optimizer with a base learning rate of 3e-4 and a batch size of 256 for 100 epochs.

Table~\ref{tab:cmp_audio} summarizes the results on the six downstream classification tasks.
Compared against BYOL-A with the Mixup augmentation, our randomized quantization outperforms it in four out of the six tasks.
Our approach is particularly stronger by a margin of 5.6\% on the VoxCeleb1 dataset, which is the hardest classification task with 1211 classes among all six tasks.
Our improvements tend to be smaller for tasks with fewer classes. 
On average, the proposed augmentation surpasses the current state-the-of-art BYOL-A by a margin of 1.8\%.


\subsection{DABS}
\label{exp_dabs}

We additionally conduct experiments on the public benchmark DABS~\cite{tamkin2021dabs} which is designed to study domain-agnostic self-supervised representation learning. Since some of its domains are discrete in nature (e.g., language), we first embed the data with a frozen layer and then augment the data embeddings for all the modalities consistently.
It contains six data modalities\footnote{The benchmark also provides an additional multi-lingual text modality. However, it is not evaluated in the original paper. We thus omit this.}, 
covering natural RGB images, multichannel sensor data, English text, audio, chest x-ray images, as well as captioned images.
In each domain, pre-training is conducted on a large-scale dataset, and the learned representations are evaluated with linear probing on various in-domain downstream datasets. The average performance for the in-domain downstream datasets is reported. We refer the reader to the benchmark for a full description of the pretraining datasets and in-domain evaluation datasets.

We follow a leading method e-Mix~\cite{tamkin2021dabs} with a Transformer architecture on this domain-agnostic benchmark.
The network is optimized with the Adam optimizer with a learning rate of 1e-4 and weight decay of 1e-4. 
The training protocol follows e-Mix, and all modalities share the same recipe.

We apply randomized quantization on the token embeddings before the Transformer. Since the quantization function has zero gradients everywhere, we randomly initialize the token embedding module without updating it. The straight-through estimator can be potentially useful, but it is not the focus of this work.  

Table~\ref{tab:cmp_dabs} summarizes the results for this benchmark. Our model outperforms the baseline e-Mix on all modalities. The improvements on natural images, speech, and sensors are larger than 3\%, while the improvements on text and chest x-rays are relatively smaller, less than 1\%. 
Both e-Mix and our pretraining seem to hurt the representation quality for captioned images. We hypothesize that the two modalities of images and texts pose significant challenges for a naive contrastive learning approach.



\section{Conclusion}

We propose randomized quantization as a novel data augmentation tool for self-supervised representation learning.
Quantization effectively withholds information within the  quantization bins but retains the information across bins.
It could be applied on arbitrary data along the channel dimension without domain-specific knowledge.
Randomized quantization significantly outperforms existing domain-agnostic augmentations based on Mixup.
It compares favorably against domain-specific augmentations on vision, and attains state-of-the-art results on audio and 3D point clouds.
We also explored randomized quantization on input data embeddings in a neural network for a wide range of data modalities. 
Experimental results on the DABS benchmark demonstrates state-of-the-art results for speech, text, images and multiple sensors. 
Randomized quantization could potentially be applied in a masked modeling framework, where the original images are reconstructed from quantized ones. This direction will be explored in future work.

\section*{Broader Impacts}

Although the proposed augmentation is generic in its formulation, it is not guaranteed to work beyond the modalities investigated in this paper. 
Application of the augmentation for other self-supervised learning frameworks such as masked modeling or
generalization to other downstream tasks remains under-explored.

\section*{Acknowledgements}
We sincerely appreciate Reviewer \#5 and AC's unwavering efforts in saving this work from rejection. This research was supported in part by National Natural Science Foundation of China/HKSAR Research Grants Council Joint Research Scheme under Grant N\_HKUST627/20, and in part by HKSAR RGC General Research Fund (GRF) \#16203319.

{\small
\bibliographystyle{ieee_fullname}
\bibliography{egbib}

\begin{thebibliography}{10}\itemsep=-1pt

\bibitem{arik2021tabnet}
Sercan~{\"O} Arik and Tomas Pfister.
\newblock Tabnet: Attentive interpretable tabular learning.
\newblock In {\em Proceedings of the AAAI Conference on Artificial
  Intelligence}, volume~35, pages 6679--6687, 2021.

\bibitem{bachman2019learning}
Philip Bachman, R~Devon Hjelm, and William Buchwalter.
\newblock Learning representations by maximizing mutual information across
  views.
\newblock {\em Advances in neural information processing systems}, 32, 2019.

\bibitem{baevski2022data2vec}
Alexei Baevski, Wei-Ning Hsu, Qiantong Xu, Arun Babu, Jiatao Gu, and Michael
  Auli.
\newblock Data2vec: A general framework for self-supervised learning in speech,
  vision and language.
\newblock {\em arXiv preprint arXiv:2202.03555}, 2022.

\bibitem{baevski2020wav2vec}
Alexei Baevski, Yuhao Zhou, Abdelrahman Mohamed, and Michael Auli.
\newblock wav2vec 2.0: A framework for self-supervised learning of speech
  representations.
\newblock {\em Advances in Neural Information Processing Systems},
  33:12449--12460, 2020.

\bibitem{banner2018scalable}
Ron Banner, Itay Hubara, Elad Hoffer, and Daniel Soudry.
\newblock Scalable methods for 8-bit training of neural networks.
\newblock {\em Advances in neural information processing systems}, 31, 2018.

\bibitem{bao2021beit}
Hangbo Bao, Li Dong, and Furu Wei.
\newblock Beit: Bert pre-training of image transformers.
\newblock {\em arXiv preprint arXiv:2106.08254}, 2021.

\bibitem{benaim2020speednet}
Sagie Benaim, Ariel Ephrat, Oran Lang, Inbar Mosseri, William~T Freeman,
  Michael Rubinstein, Michal Irani, and Tali Dekel.
\newblock Speednet: Learning the speediness in videos.
\newblock In {\em Proceedings of the IEEE/CVF Conference on Computer Vision and
  Pattern Recognition}, pages 9922--9931, 2020.

\bibitem{bengio2013estimating}
Yoshua Bengio, Nicholas L{\'e}onard, and Aaron Courville.
\newblock Estimating or propagating gradients through stochastic neurons for
  conditional computation.
\newblock {\em arXiv preprint arXiv:1308.3432}, 2013.

\bibitem{bommasani2021opportunities}
Rishi Bommasani, Drew~A Hudson, Ehsan Adeli, Russ Altman, Simran Arora, Sydney
  von Arx, Michael~S Bernstein, Jeannette Bohg, Antoine Bosselut, Emma
  Brunskill, et~al.
\newblock On the opportunities and risks of foundation models.
\newblock {\em arXiv preprint arXiv:2108.07258}, 2021.

\bibitem{bowles2018gan}
Christopher Bowles, Liang Chen, Ricardo Guerrero, Paul Bentley, Roger Gunn,
  Alexander Hammers, David~Alexander Dickie, Maria~Vald{\'e}s Hern{\'a}ndez,
  Joanna Wardlaw, and Daniel Rueckert.
\newblock Gan augmentation: Augmenting training data using generative
  adversarial networks.
\newblock {\em arXiv preprint arXiv:1810.10863}, 2018.

\bibitem{brislin1970back}
Richard~W Brislin.
\newblock Back-translation for cross-cultural research.
\newblock {\em Journal of cross-cultural psychology}, 1(3):185--216, 1970.

\bibitem{cao2022synergistic}
Yun-Hao Cao, Peiqin Sun, Yechang Huang, Jianxin Wu, and Shuchang Zhou.
\newblock Synergistic self-supervised and quantization learning.
\newblock In {\em ECCV}, 2022.

\bibitem{caron2021emerging}
Mathilde Caron, Hugo Touvron, Ishan Misra, Herv{\'e} J{\'e}gou, Julien Mairal,
  Piotr Bojanowski, and Armand Joulin.
\newblock Emerging properties in self-supervised vision transformers.
\newblock In {\em Proceedings of the IEEE/CVF International Conference on
  Computer Vision}, pages 9650--9660, 2021.

\bibitem{chang2015shapenet}
Angel~X Chang, Thomas Funkhouser, Leonidas Guibas, Pat Hanrahan, Qixing Huang,
  Zimo Li, Silvio Savarese, Manolis Savva, Shuran Song, Hao Su, et~al.
\newblock Shapenet: An information-rich 3d model repository.
\newblock {\em arXiv preprint arXiv:1512.03012}, 2015.

\bibitem{chen2020statistical}
Jianfei Chen, Yu Gai, Zhewei Yao, Michael~W Mahoney, and Joseph~E Gonzalez.
\newblock A statistical framework for low-bitwidth training of deep neural
  networks.
\newblock {\em Advances in Neural Information Processing Systems}, 33:883--894,
  2020.

\bibitem{chen2021empirical}
Xinlei Chen, Saining Xie, and Kaiming He.
\newblock An empirical study of training self-supervised vision transformers.
\newblock In {\em Proceedings of the IEEE/CVF International Conference on
  Computer Vision}, pages 9640--9649, 2021.

\bibitem{clark2020electra}
Kevin Clark, Minh-Thang Luong, Quoc~V Le, and Christopher~D Manning.
\newblock Electra: Pre-training text encoders as discriminators rather than
  generators.
\newblock {\em arXiv preprint arXiv:2003.10555}, 2020.

\bibitem{courbariaux2014training}
Matthieu Courbariaux, Yoshua Bengio, and Jean-Pierre David.
\newblock Training deep neural networks with low precision multiplications.
\newblock {\em arXiv preprint arXiv:1412.7024}, 2014.

\bibitem{courbariaux2015binaryconnect}
Matthieu Courbariaux, Yoshua Bengio, and Jean-Pierre David.
\newblock Binaryconnect: Training deep neural networks with binary weights
  during propagations.
\newblock {\em Advances in neural information processing systems}, 28, 2015.

\bibitem{cramer2019look}
Jason Cramer, Ho-Hsiang Wu, Justin Salamon, and Juan~Pablo Bello.
\newblock Look, listen, and learn more: Design choices for deep audio
  embeddings.
\newblock In {\em ICASSP 2019-2019 IEEE International Conference on Acoustics,
  Speech and Signal Processing (ICASSP)}, pages 3852--3856. IEEE, 2019.

\bibitem{cubuk2018autoaugment}
Ekin~D Cubuk, Barret Zoph, Dandelion Mane, Vijay Vasudevan, and Quoc~V Le.
\newblock Autoaugment: Learning augmentation policies from data.
\newblock {\em arXiv preprint arXiv:1805.09501}, 2018.

\bibitem{deng2009imagenet}
Jia Deng, Wei Dong, Richard Socher, Li-Jia Li, Kai Li, and Li Fei-Fei.
\newblock Imagenet: A large-scale hierarchical image database.
\newblock In {\em 2009 IEEE conference on computer vision and pattern
  recognition}, pages 248--255. Ieee, 2009.

\bibitem{devlin2018bert}
Jacob Devlin, Ming-Wei Chang, Kenton Lee, and Kristina Toutanova.
\newblock Bert: Pre-training of deep bidirectional transformers for language
  understanding.
\newblock {\em arXiv preprint arXiv:1810.04805}, 2018.

\bibitem{devries2017improved}
Terrance DeVries and Graham~W Taylor.
\newblock Improved regularization of convolutional neural networks with cutout.
\newblock {\em arXiv preprint arXiv:1708.04552}, 2017.

\bibitem{doersch2015unsupervised}
Carl Doersch, Abhinav Gupta, and Alexei~A Efros.
\newblock Unsupervised visual representation learning by context prediction.
\newblock In {\em Proceedings of the IEEE international conference on computer
  vision}, pages 1422--1430, 2015.

\bibitem{dwibedi2021little}
Debidatta Dwibedi, Yusuf Aytar, Jonathan Tompson, Pierre Sermanet, and Andrew
  Zisserman.
\newblock With a little help from my friends: Nearest-neighbor contrastive
  learning of visual representations.
\newblock In {\em Proceedings of the IEEE/CVF International Conference on
  Computer Vision}, pages 9588--9597, 2021.

\bibitem{engel2017neural}
Jesse Engel, Cinjon Resnick, Adam Roberts, Sander Dieleman, Mohammad Norouzi,
  Douglas Eck, and Karen Simonyan.
\newblock Neural audio synthesis of musical notes with wavenet autoencoders.
\newblock In {\em International Conference on Machine Learning}, pages
  1068--1077. PMLR, 2017.

\bibitem{fan2020training}
Angela Fan, Pierre Stock, Benjamin Graham, Edouard Grave, R{\'e}mi Gribonval,
  Herve Jegou, and Armand Joulin.
\newblock Training with quantization noise for extreme model compression.
\newblock {\em arXiv preprint arXiv:2004.07320}, 2020.

\bibitem{favory2020coala}
Xavier Favory, Konstantinos Drossos, Tuomas Virtanen, and Xavier Serra.
\newblock Coala: Co-aligned autoencoders for learning semantically enriched
  audio representations.
\newblock {\em arXiv preprint arXiv:2006.08386}, 2020.

\bibitem{fu2022contrastive}
Yonggan Fu, Qixuan Yu, Meng Li, Xu Ouyang, Vikas Chandra, and Yingyan Lin.
\newblock Contrastive quant: quantization makes stronger contrastive learning.
\newblock In {\em Proceedings of the 59th ACM/IEEE Design Automation
  Conference}, pages 205--210, 2022.

\bibitem{gemmeke2017audio}
Jort~F Gemmeke, Daniel~PW Ellis, Dylan Freedman, Aren Jansen, Wade Lawrence,
  R~Channing Moore, Manoj Plakal, and Marvin Ritter.
\newblock Audio set: An ontology and human-labeled dataset for audio events.
\newblock In {\em 2017 IEEE international conference on acoustics, speech and
  signal processing (ICASSP)}, pages 776--780. IEEE, 2017.

\bibitem{gholami2021survey}
Amir Gholami, Sehoon Kim, Zhen Dong, Zhewei Yao, Michael~W Mahoney, and Kurt
  Keutzer.
\newblock A survey of quantization methods for efficient neural network
  inference.
\newblock {\em arXiv preprint arXiv:2103.13630}, 2021.

\bibitem{gray1998quantization}
Robert~M. Gray and David~L. Neuhoff.
\newblock Quantization.
\newblock {\em IEEE transactions on information theory}, 44(6):2325--2383,
  1998.

\bibitem{grill2020bootstrap}
Jean-Bastien Grill, Florian Strub, Florent Altch{\'e}, Corentin Tallec, Pierre
  Richemond, Elena Buchatskaya, Carl Doersch, Bernardo Avila~Pires, Zhaohan
  Guo, Mohammad Gheshlaghi~Azar, et~al.
\newblock Bootstrap your own latent-a new approach to self-supervised learning.
\newblock {\em Advances in neural information processing systems},
  33:21271--21284, 2020.

\bibitem{guo2020nonlinear}
Hongyu Guo.
\newblock Nonlinear mixup: Out-of-manifold data augmentation for text
  classification.
\newblock In {\em Proceedings of the AAAI Conference on Artificial
  Intelligence}, volume~34, pages 4044--4051, 2020.

\bibitem{gupta2015deep}
Suyog Gupta, Ankur Agrawal, Kailash Gopalakrishnan, and Pritish Narayanan.
\newblock Deep learning with limited numerical precision.
\newblock In {\em International conference on machine learning}, pages
  1737--1746. PMLR, 2015.

\bibitem{Hassani2019}
Kaveh Hassani and Mike Haley.
\newblock Unsupervised multi-task feature learning on point clouds.
\newblock In {\em ICCV}, 2019.

\bibitem{he2022masked}
Kaiming He, Xinlei Chen, Saining Xie, Yanghao Li, Piotr Doll{\'a}r, and Ross
  Girshick.
\newblock Masked autoencoders are scalable vision learners.
\newblock In {\em Proceedings of the IEEE/CVF Conference on Computer Vision and
  Pattern Recognition}, pages 16000--16009, 2022.

\bibitem{he2016deep}
Kaiming He, Xiangyu Zhang, Shaoqing Ren, and Jian Sun.
\newblock Deep residual learning for image recognition.
\newblock In {\em Proceedings of the IEEE conference on computer vision and
  pattern recognition}, pages 770--778, 2016.

\bibitem{hendrycks2019augmix}
Dan Hendrycks, Norman Mu, Ekin~D Cubuk, Barret Zoph, Justin Gilmer, and Balaji
  Lakshminarayanan.
\newblock Augmix: A simple data processing method to improve robustness and
  uncertainty.
\newblock {\em arXiv preprint arXiv:1912.02781}, 2019.

\bibitem{hinton2009deep}
Geoffrey~E Hinton.
\newblock Deep belief networks.
\newblock {\em Scholarpedia}, 4(5):5947, 2009.

\bibitem{hinton1993autoencoders}
Geoffrey~E Hinton and Richard Zemel.
\newblock Autoencoders, minimum description length and helmholtz free energy.
\newblock {\em Advances in neural information processing systems}, 6, 1993.

\bibitem{huang2021ascnet}
Deng Huang, Wenhao Wu, Weiwen Hu, Xu Liu, Dongliang He, Zhihua Wu, Xiangmiao
  Wu, Mingkui Tan, and Errui Ding.
\newblock Ascnet: Self-supervised video representation learning with
  appearance-speed consistency.
\newblock In {\em Proceedings of the IEEE/CVF International Conference on
  Computer Vision}, pages 8096--8105, 2021.

\bibitem{huffman1952method}
David~A Huffman.
\newblock A method for the construction of minimum-redundancy codes.
\newblock {\em Proceedings of the IRE}, 40(9):1098--1101, 1952.

\bibitem{jernite2017discourse}
Yacine Jernite, Samuel~R Bowman, and David Sontag.
\newblock Discourse-based objectives for fast unsupervised sentence
  representation learning.
\newblock {\em arXiv preprint arXiv:1705.00557}, 2017.

\bibitem{kiros2015skip}
Ryan Kiros, Yukun Zhu, Russ~R Salakhutdinov, Richard Zemel, Raquel Urtasun,
  Antonio Torralba, and Sanja Fidler.
\newblock Skip-thought vectors.
\newblock {\em Advances in neural information processing systems}, 28, 2015.

\bibitem{koohpayegani2021mean}
Soroush~Abbasi Koohpayegani, Ajinkya Tejankar, and Hamed Pirsiavash.
\newblock Mean shift for self-supervised learning.
\newblock In {\em Proceedings of the IEEE/CVF International Conference on
  Computer Vision}, pages 10326--10335, 2021.

\bibitem{lee2020mix}
Kibok Lee, Yian Zhu, Kihyuk Sohn, Chun-Liang Li, Jinwoo Shin, and Honglak Lee.
\newblock i-mix: A domain-agnostic strategy for contrastive representation
  learning.
\newblock {\em arXiv preprint arXiv:2010.08887}, 2020.

\bibitem{lin2017towards}
Xiaofan Lin, Cong Zhao, and Wei Pan.
\newblock Towards accurate binary convolutional neural network.
\newblock {\em Advances in neural information processing systems}, 30, 2017.

\bibitem{lucas2018mixed}
Thomas Lucas, Corentin Tallec, Yann Ollivier, and Jakob Verbeek.
\newblock Mixed batches and symmetric discriminators for gan training.
\newblock In {\em International Conference on Machine Learning}, pages
  2844--2853. PMLR, 2018.

\bibitem{micikevicius2017mixed}
Paulius Micikevicius, Sharan Narang, Jonah Alben, Gregory Diamos, Erich Elsen,
  David Garcia, Boris Ginsburg, Michael Houston, Oleksii Kuchaiev, Ganesh
  Venkatesh, et~al.
\newblock Mixed precision training.
\newblock {\em arXiv preprint arXiv:1710.03740}, 2017.

\bibitem{misra2016shuffle}
Ishan Misra, C~Lawrence Zitnick, and Martial Hebert.
\newblock Shuffle and learn: unsupervised learning using temporal order
  verification.
\newblock In {\em European conference on computer vision}, pages 527--544.
  Springer, 2016.

\bibitem{nagrani2017voxceleb}
Arsha Nagrani, Joon~Son Chung, and Andrew Zisserman.
\newblock Voxceleb: a large-scale speaker identification dataset.
\newblock {\em arXiv preprint arXiv:1706.08612}, 2017.

\bibitem{niizumi2021byol}
Daisuke Niizumi, Daiki Takeuchi, Yasunori Ohishi, Noboru Harada, and Kunio
  Kashino.
\newblock Byol for audio: Self-supervised learning for general-purpose audio
  representation.
\newblock In {\em 2021 International Joint Conference on Neural Networks
  (IJCNN)}, pages 1--8. IEEE, 2021.

\bibitem{noroozi2016unsupervised}
Mehdi Noroozi and Paolo Favaro.
\newblock Unsupervised learning of visual representations by solving jigsaw
  puzzles.
\newblock In {\em European conference on computer vision}, pages 69--84.
  Springer, 2016.

\bibitem{oord2018representation}
Aaron van~den Oord, Yazhe Li, and Oriol Vinyals.
\newblock Representation learning with contrastive predictive coding.
\newblock {\em arXiv preprint arXiv:1807.03748}, 2018.

\bibitem{park2019specaugment}
Daniel~S Park, William Chan, Yu Zhang, Chung-Cheng Chiu, Barret Zoph, Ekin~D
  Cubuk, and Quoc~V Le.
\newblock Specaugment: A simple data augmentation method for automatic speech
  recognition.
\newblock {\em arXiv preprint arXiv:1904.08779}, 2019.

\bibitem{pathak2016context}
Deepak Pathak, Philipp Krahenbuhl, Jeff Donahue, Trevor Darrell, and Alexei~A
  Efros.
\newblock Context encoders: Feature learning by inpainting.
\newblock In {\em Proceedings of the IEEE conference on computer vision and
  pattern recognition}, pages 2536--2544, 2016.

\bibitem{saeed2021contrastive}
Aaqib Saeed, David Grangier, and Neil Zeghidour.
\newblock Contrastive learning of general-purpose audio representations.
\newblock In {\em ICASSP 2021-2021 IEEE International Conference on Acoustics,
  Speech and Signal Processing (ICASSP)}, pages 3875--3879. IEEE, 2021.

\bibitem{salamon2014dataset}
Justin Salamon, Christopher Jacoby, and Juan~Pablo Bello.
\newblock A dataset and taxonomy for urban sound research.
\newblock In {\em Proceedings of the 22nd ACM international conference on
  Multimedia}, pages 1041--1044, 2014.

\bibitem{shannon1948mathematical}
Claude~Elwood Shannon.
\newblock A mathematical theory of communication.
\newblock {\em The Bell system technical journal}, 27(3):379--423, 1948.

\bibitem{shannon1959coding}
Claude~E Shannon et~al.
\newblock Coding theorems for a discrete source with a fidelity criterion.
\newblock {\em IRE Nat. Conv. Rec}, 4(142-163):1, 1959.

\bibitem{sheppard1897calculation}
William~Fleetwood Sheppard.
\newblock On the calculation of the most probable values of
  frequency-constants, for data arranged according to equidistant division of a
  scale.
\newblock {\em Proceedings of the London Mathematical Society}, 1(1):353--380,
  1897.

\bibitem{shor2020towards}
Joel Shor, Aren Jansen, Ronnie Maor, Oran Lang, Omry Tuval, Felix de~Chaumont
  Quitry, Marco Tagliasacchi, Ira Shavitt, Dotan Emanuel, and Yinnon Haviv.
\newblock Towards learning a universal non-semantic representation of speech.
\newblock {\em arXiv preprint arXiv:2002.12764}, 2020.

\bibitem{sun2019infograph}
Fan-Yun Sun, Jordan Hoffmann, Vikas Verma, and Jian Tang.
\newblock Infograph: Unsupervised and semi-supervised graph-level
  representation learning via mutual information maximization.
\newblock {\em arXiv preprint arXiv:1908.01000}, 2019.

\bibitem{tamkin2021dabs}
Alex Tamkin, Vincent Liu, Rongfei Lu, Daniel Fein, Colin Schultz, and Noah
  Goodman.
\newblock Dabs: A domain-agnostic benchmark for self-supervised learning.
\newblock {\em arXiv preprint arXiv:2111.12062}, 2021.

\bibitem{tong2022videomae}
Zhan Tong, Yibing Song, Jue Wang, and Limin Wang.
\newblock Videomae: Masked autoencoders are data-efficient learners for
  self-supervised video pre-training.
\newblock {\em arXiv preprint arXiv:2203.12602}, 2022.

\bibitem{verma2021towards}
Vikas Verma, Thang Luong, Kenji Kawaguchi, Hieu Pham, and Quoc Le.
\newblock Towards domain-agnostic contrastive learning.
\newblock In {\em International Conference on Machine Learning}, pages
  10530--10541. PMLR, 2021.

\bibitem{wang2018training}
Naigang Wang, Jungwook Choi, Daniel Brand, Chia-Yu Chen, and Kailash
  Gopalakrishnan.
\newblock Training deep neural networks with 8-bit floating point numbers.
\newblock {\em Advances in neural information processing systems}, 31, 2018.

\bibitem{wang2017cnn}
Peng-Shuai Wang, Yang Liu, Yu-Xiao Guo, Chun-Yu Sun, and Xin Tong.
\newblock O-cnn: Octree-based convolutional neural networks for 3d shape
  analysis.
\newblock {\em ACM Transactions On Graphics (TOG)}, 36(4):1--11, 2017.

\bibitem{wang2021unsupervised}
Peng-Shuai Wang, Yu-Qi Yang, Qian-Fang Zou, Zhirong Wu, Yang Liu, and Xin Tong.
\newblock Unsupervised 3d learning for shape analysis via multiresolution
  instance discrimination.
\newblock In {\em Proceedings of the AAAI Conference on Artificial
  Intelligence}, volume~35, pages 2773--2781, 2021.

\bibitem{warden2018speech}
Pete Warden.
\newblock Speech commands: A dataset for limited-vocabulary speech recognition.
\newblock {\em arXiv preprint arXiv:1804.03209}, 2018.

\bibitem{wei2018learning}
Donglai Wei, Joseph~J Lim, Andrew Zisserman, and William~T Freeman.
\newblock Learning and using the arrow of time.
\newblock In {\em Proceedings of the IEEE Conference on Computer Vision and
  Pattern Recognition}, pages 8052--8060, 2018.

\bibitem{wei2019eda}
Jason Wei and Kai Zou.
\newblock Eda: Easy data augmentation techniques for boosting performance on
  text classification tasks.
\newblock {\em arXiv preprint arXiv:1901.11196}, 2019.

\bibitem{wu2022extreme}
Zhirong Wu, Zihang Lai, Xiao Sun, and Stephen Lin.
\newblock Extreme masking for learning instance and distributed visual
  representations.
\newblock {\em arXiv preprint arXiv:2206.04667}, 2022.

\bibitem{wu2015adjustable}
Zhirong Wu, Dahua Lin, and Xiaoou Tang.
\newblock Adjustable bounded rectifiers: Towards deep binary representations.
\newblock {\em arXiv preprint arXiv:1511.06201}, 2015.

\bibitem{wu20153d}
Zhirong Wu, Shuran Song, Aditya Khosla, Fisher Yu, Linguang Zhang, Xiaoou Tang,
  and Jianxiong Xiao.
\newblock 3d shapenets: A deep representation for volumetric shapes.
\newblock In {\em Proceedings of the IEEE conference on computer vision and
  pattern recognition}, pages 1912--1920, 2015.

\bibitem{wu2018unsupervised}
Zhirong Wu, Yuanjun Xiong, Stella~X Yu, and Dahua Lin.
\newblock Unsupervised feature learning via non-parametric instance
  discrimination.
\newblock In {\em Proceedings of the IEEE conference on computer vision and
  pattern recognition}, pages 3733--3742, 2018.

\bibitem{xu2022masked}
Hu Xu, Juncheng Li, Alexei Baevski, Michael Auli, Wojciech Galuba, Florian
  Metze, Christoph Feichtenhofer, et~al.
\newblock Masked autoencoders that listen.
\newblock {\em arXiv preprint arXiv:2207.06405}, 2022.

\bibitem{Yang2018a}
Yaoqing Yang, Chen Feng, Yiru Shen, and Dong Tian.
\newblock {FoldingNet}: Point cloud auto-encoder via deep grid deformation.
\newblock In {\em CVPR}, 2018.

\bibitem{yi2017large}
Li Yi, Lin Shao, Manolis Savva, Haibin Huang, Yang Zhou, Qirui Wang, Benjamin
  Graham, Martin Engelcke, Roman Klokov, Victor Lempitsky, et~al.
\newblock Large-scale 3d shape reconstruction and segmentation from shapenet
  core55.
\newblock {\em arXiv preprint arXiv:1710.06104}, 2017.

\bibitem{yun2019cutmix}
Sangdoo Yun, Dongyoon Han, Seong~Joon Oh, Sanghyuk Chun, Junsuk Choe, and
  Youngjoon Yoo.
\newblock Cutmix: Regularization strategy to train strong classifiers with
  localizable features.
\newblock In {\em Proceedings of the IEEE/CVF international conference on
  computer vision}, pages 6023--6032, 2019.

\bibitem{zbontar2021barlow}
Jure Zbontar, Li Jing, Ishan Misra, Yann LeCun, and St{\'e}phane Deny.
\newblock Barlow twins: Self-supervised learning via redundancy reduction.
\newblock In {\em International Conference on Machine Learning}, pages
  12310--12320. PMLR, 2021.

\bibitem{zhang2021understanding}
Chiyuan Zhang, Samy Bengio, Moritz Hardt, Benjamin Recht, and Oriol Vinyals.
\newblock Understanding deep learning (still) requires rethinking
  generalization.
\newblock {\em Communications of the ACM}, 64(3):107--115, 2021.

\bibitem{zhang2017mixup}
Hongyi Zhang, Moustapha Cisse, Yann~N Dauphin, and David Lopez-Paz.
\newblock mixup: Beyond empirical risk minimization.
\newblock {\em arXiv preprint arXiv:1710.09412}, 2017.

\bibitem{zhang2016colorful}
Richard Zhang, Phillip Isola, and Alexei~A Efros.
\newblock Colorful image colorization.
\newblock In {\em European conference on computer vision}, pages 649--666.
  Springer, 2016.

\bibitem{zhao2021distilling}
Nanxuan Zhao, Zhirong Wu, Rynson~WH Lau, and Stephen Lin.
\newblock Distilling localization for self-supervised representation learning.
\newblock In {\em Proceedings of the AAAI Conference on Artificial
  Intelligence}, volume~35, pages 10990--10998, 2021.

\end{thebibliography}
}

\end{document}